\documentclass{article}

\usepackage{arxiv}

\usepackage[utf8]{inputenc} % allow utf-8 input
\usepackage[T1]{fontenc}    % use 8-bit T1 fonts
\usepackage{hyperref}       % hyperlinks
\usepackage{url}            % simple URL typesetting
\usepackage{booktabs}       % professional-quality tables
\usepackage{amsfonts}       % blackboard math symbols
\usepackage{nicefrac}       % compact symbols for 1/2, etc.
\usepackage{microtype}      % microtypography
\usepackage{lipsum}
\usepackage{graphicx}
\graphicspath{ {./images/} }

\usepackage{amsmath,amssymb,amsfonts}
\usepackage{bbm}
\usepackage{algorithmic}
\usepackage{graphicx}
\usepackage{textcomp}

\usepackage{listings} % for code snippets

\usepackage[numbers]{natbib}

\usepackage{import}
\usepackage{booktabs}
\usepackage{xcolor}
\usepackage{colortbl}
\usepackage{rotating}
\usepackage{multirow}
\usepackage[normalem]{ulem} % for underline in table, the option is to fix 

\title{TEE4EHR: Transformer Event Encoder for Better Representation Learning in Electronic Health Records}

\author{
 Hojjat Karami \\
  EPFL \\
  \texttt{hojjat.karami@epfl.ch} \\
   \And
 David Atienza \\
  EPFL\\
  \And
 Anisoara Ionescu \\
  EPFL\\
}

\begin{document}
\maketitle
\begin{abstract}
Irregular sampling of time series in electronic health records (EHRs) is one of the main challenges for developing machine learning models. Additionally, the pattern of missing data in certain clinical variables is not at random but depends on the decisions of clinicians and the state of the patient. Point process is a mathematical framework for analyzing event sequence data that is consistent with irregular sampling patterns. Our model, \emph{TEE4EHR}, is a transformer event encoder (TEE) with point process loss that encodes the pattern of laboratory tests in EHRs. The utility of our TEE has been investigated in a variety of benchmark event sequence datasets. Additionally, we conduct experiments on two real-world EHR databases to provide a more comprehensive evaluation of our model. Firstly, in a self-supervised learning approach, the TEE is jointly learned with an existing attention-based deep neural network which gives superior performance in negative log-likelihood and future event prediction. Besides, we propose an algorithm for aggregating attention weights that can reveal the interaction between the events. Secondly, we transfer and freeze the learned TEE to the downstream task for the outcome prediction, where it outperforms state-of-the-art models for handling irregularly sampled time series. Furthermore, our results demonstrate that our approach can improve representation learning in EHRs and can be useful for clinical prediction tasks.\footnote{The code will be available on github after acceptance. for more information, please email the corresponding author.}
\end{abstract}

% keywords can be removed

\section{Introduction}

Machine learning has the potential to revolutionize healthcare by leveraging the vast amounts of data available in electronic health records (EHRs) to develop more accurate clinical decision support systems \cite{peiffer-smadjaMachineLearningClinical2020,pourhomayounPredictingMortalityRisk2021}. EHRs store patient health information, such as medical history, medications, lab results, and diagnostic images, which can be used as input for machine learning algorithms to identify patterns and associations that could inform more precise diagnoses \cite{kaoContextAwareSymptomChecking2018,calvertMachineLearningBasedLaboratoryDeveloped2019}, better treatment plans \cite{zhangLEAPLearningPrescribe2017} , and earlier interventions \cite{batesReportingImplementingInterventions2020}. Clinical decision support systems that use machine learning can provide evidence-based real-time recommendations to healthcare providers, reducing errors, and improving patient outcomes \cite{lopez-martinezCaseStudyBig2020}.

Irregular sampling is one of the data challenges for machine learning (ML) when using electronic health records (EHRs). EHR data is often collected at different times and frequencies, depending on a patient's healthcare needs and visit schedules, which can result in uneven and irregularly sampled time series data. From a data perspective, asynchronous and incomplete observation of certain clinical variables is regarded as missingness in the data. However, the sources of missing data in EHRs must be carefully understood. For instance, lab measurements are usually ordered as part of a routine care or diagnostic workup, and the presence or absence of a data point conveys information about the patient's state \cite{ghassemiReviewChallengesOpportunities2020}. As a result, in most cases, the missingness is not at random (MNAR) and must be handled with care. In this paper, we refer to this type of missingness as \emph{informative missingness}.

The most straightforward solution to handle missing data in EHR time series is to use imputation techniques. Imputation refers to the process of filling in the missing data with plausible values based on the available information. One simple approach is to impute with mean, mode (most frequent), median, or more sophisticated machine learning-based imputation methods such as MissForest \cite{stekhovenMissForestNonparametricMissing2012}, MICE \cite{buurenMiceMultivariateImputation2011}, and kNN imputer \cite{duyleComparisonMostInfluential2018} have been developed that can better capture complex relationships in the data. Regardless of the technique used, it is important to carefully consider the impact of the imputed values on the analysis and report the imputation method used in the results. Besides, the amount of missing data in EHRs is often large, and imputation can be computationally expensive. In addition, the occurrence or non-occurrence of a measurement and how often it is observed can convey its own information and imputation may lead to an undesired distribution shift \cite{zhangGraphGuidedNetworkIrregularly2022}. Therefore, filling in the missing values may not always be preferable \cite{littleStatisticalAnalysisMissing2019}.

Recently, new machine learning models have been developed to handle irregularly sampled time series without imputation. Gaussian Process (GP) can handle missing data effectively by providing a coherent framework for imputing missing values while capturing the uncertainty associated with the imputations \cite{ghassemiDatadrivenApproachOptimized2014}. Deep learning models have also been employed to handle irregularly sampled time series. Recurrent Neural Networks (RNNs) are well-suited for this task because they can process sequential data, taking into account not only the current input but also the previous inputs. Convolutional Neural Networks (CNNs) can also extract features from the data and then pass the processed data to an RNN for further analysis \cite{DCSFDeepConvolutional,kosmaTimeParameterizedConvolutionalNeural2023}. Additionally, Attention weights can be integrated into RNNs and CNNs to help the model focus on important features in the data \cite{zhangAttentionBasedConvolutional2021,niuReviewAttentionMechanism2021}. These deep learning models offer a powerful toolset for handling irregularly sampled time series and providing useful insights into this type of data. However, it is important to note that these models do not explicitly account for the fact that the missing data can convey unique information.

Point process is a mathematical framework for describing the distribution of events in time or space. These events can be user activities on social media \cite{rizoiuHawkesProcessesEvents2017}, financial transactions \cite{bauwensModellingFinancialHigh2009}, gene positions in bioinformatics \cite{reynaud-bouretAdaptiveEstimationHawkes2010}, or even the pattern of laboratory tests in EHRs that can be regarded as a sequence of events ordered by clinicians \cite{alaaLearningClinicalJudgments2017}. At the core of the point process are the definition of conditional intensity functions (CIFs) and the corresponding log-likelihood, which simultaneously models the occurrence of events using the history of past events.
More recently, Neural Point Processes (NPPs) have been developed to better characterize CIFs by leveraging the power of deep neural networks \cite{shchurNeuralTemporalPoint2021}. These models can be used for tasks such as predicting future events, estimating the rate of event occurrence, or identifying correlations between events \cite{cheRecurrentNeuralNetworks2018,zhangSelfAttentiveHawkesProcess2020, zuoTransformerHawkesProcess2020}. They offer a flexible and powerful way to analyze event sequence data, as they can handle complex dependencies between events and incorporate prior knowledge about the process.

\subsection*{Aim of Study}

Our primary objective is to enhance the capability of deep learning models for irregularly sampled time series by leveraging the capabilities of neural point processes. We propose a new framework, \emph{TEE4EHR}, which is a Transformer Event Encoder (TEE) with a Deep Attention Module (DAM) for irregularly sampled time series in electronic health records (EHRs). Our TEE is inspired by attention-based neural point processes \cite{zhangSelfAttentiveHawkesProcess2020, zuoTransformerHawkesProcess2020} which regards the pattern of irregularly sampled time series as a sequence of events and can be combined with any existing deep learning model for irregularly sampled time series. The code is available at \url{https://github.com/hojjatkarami/TEE4EHR}.

% In the first part, we show the improved performance of our TEE by conducting various experiments on common event sequence datasets such as Stack Overflow and ReTweets. In the second part, we conduct various experiments on two real-world longitudinal healthcare datasets from PhysioNet challenges \cite{silvaPredictingInHospitalMortality, reynaEarlyPredictionSepsis2020}. In this case, TEE will only take into account the pattern of laboratory tests and is jointly trained with a recent deep attention module (DAM) \cite{hornSetFunctionsTime2020} that is particularly developed for handling irregularly sampled time series without any imputation. The training procedure is as follows: First, we train TEE jointly with DAM with different point process-based loss functions (self-supervised learning); Second, we transfer and freeze TEE to a downstream task for binary outcome prediction. We also compare our approach with state-of-the-art deep learning models for irregularly sampled time series.

\subsection*{Contributions of this work}
Our main contributions can be summarized as follows:
\begin{itemize}
  \item A new transformer event encoder (TEE) for learning CIFs and future event prediction that can be trained with different point-process loss functions.
  \item Our TEE can improve the performance metrics on benchmark datasets in neural point process literature.
  \item We present a new framework, TEE4EHR, for learning TEE jointly with an existing deep learning model that is compatible with irregularly sampled time series.
  \item We show the utility of the proposed approach on two real-world datasets, as evidenced by its superior performance in clinical outcome prediction as well as better representation learning.
\end{itemize}

The current work is an extension of the previous work presented at the 2023 IEEE-EMBS International Conference on Biomedical and Health Informatics (BHI'23). It provides a comprehensive analysis of the proposed transformer event encoder on benchmark datasets and a more in-depth explanation of the results.

\section{Background \& Related Works}

\subsection{Problem Formulation}

An irregularly sampled data can be denoted as $\mathcal{D}=\{\mathcal{U}_i\}_{i=1}^{N}$ where $N$ is the number of samples. Each sample is represented as a sequence of tuples $\mathcal{U}_i=\{(t_p,k_p,v_p)\}_{p=1}^{P_i}$ where $P_i$ is the total number of observations and $t_p, k_p, v_p$ represents the time, name, and value of $p$-th observation, respectively.

An event sequence data can be represented as $\mathcal{D}=\{\mathcal{S}_{i}\}_{i=1}^N$ where $N$ is the total number of samples. Each sample $\mathcal{S}_{i}$ is represented as a sequence of events $\mathcal{S}_{i}=\{(t_j,e_j)\}_{j=1}^{L_{i}}$, where $L_i$ is the total number of occurred events, $t_j$ is the event's timestamp, and $e_j \in \mathbb{R}^M $ is the binary representation of event marks (one-hot for multi-class or multi-hot for multi-label). Furthermore, the history of events at time $t$ is denoted as $\mathcal{H}_t=\{(t_j,e_j):t_j<t \}$.

We can represent an EHR dataset as  $\mathcal{D}=\{(\mathcal{S}_i,\mathcal{U}_i,d_i,y_i)\}_{i=1}^{N}$, where $N$ is the number of patients, $\mathcal{U}_i$ is irregularly sampled time series data, and $\mathcal{S}_{i}$ is the patient's event data such as visits and laboratory tests. Static variables such as demographics and patient's outcome, if available, are denoted by $d_i$ and $y_i$, respectively. In this work, we consider sampling pattern of time series as events as they are informative.

\subsection{Point Process Framework}

% \subsubsection*{Point Process Framework}

The core idea of the point process framework is the definition of conditional intensity functions (CIFs) which is the probability of the occurrence of an event of type $m$ in an infinitesimal time window $[t,t+dt)$:

        \begin{equation} \label{eq:lambda_def}
          \lambda_m^{*}(t)=\lim \frac{P(  \text{event of type m in } [t,t+\Delta t) |   \mathcal{H}_t  )}{\Delta t}.
        \end{equation}

        Here, $*$ denotes conditioning on the history of events ($\mathcal{H}_t$). The Multivariate Hawkes process is the traditional approach to characterize CIFs by assuming a fixed form of intensity to account for the additive influence of an event in the past:

        \begin{equation*}
          \lambda_m^{*}(t)= \mu_m + \sum_{(t^{\prime},e^{\prime})\in \mathcal{H}_t} \phi(t-t^{\prime}),
        \end{equation*}

        % where $\mu \ge 0 $, a.k.a. base intensity, is an exogenous component that is independent of history, while $\phi(t)>0$, excitation function, is an endogenous component depending on the history that shows the mutual influences \cite{zhangSelfAttentiveHawkesProcess2020}. The excitation function can be characterized using different approaches such as exponentials \cite{hawkesSpectraSelfexcitingMutually1971}, or a linear combination of $M$ basis functions \cite{xuLearningGrangerCausality2016}.

        where $\mu \ge 0$ (base intensity) is an exogenous component that is independent of history, while $\phi(t) > 0$ (excitation function) is an endogenous component depending on the history that shows the mutual influences \cite{zhangSelfAttentiveHawkesProcess2020}. The excitation function can be characterized using different approaches such as exponentials \cite{hawkesSpectraSelfexcitingMutually1971}, or a linear combination of $M$ basis functions \cite{xuLearningGrangerCausality2016}.

        \subsubsection*{Parameter Estimation}

        Based on the conditional intensity function equation (\ref{eq:lambda_def}), it is straightforward to derive conditional probability density function $p^{*}_{m}(t)$ in the interval $(t_j, t_{j+1}]$:

\begin{equation*}
  p_m^{*}(t)=\lambda_m^{*}(t) \exp \left[-\sum_{m=1}^{M} \int_{t_j}^{t_{j+1}}\lambda_m^{*}(s)ds\right]
\end{equation*}

The parameters of the point process can be learned using Maximum Likelihood Estimation (MLE) framework. However, more advanced methods such as adversarial learning \cite{yanImprovingMaximumLikelihood2018}, and reinforcement learning have also been proposed \cite{DeepReinforcementLearninga,raghuDeepReinforcementLearning2017}.

In the multi-class setting (MC) where only one event can occur at a time, the log-likelihood (LL) of the point process for a single event sequence $\mathcal{S}_{i}$ is defined as:

\begin{multline} \label{eq:CIF-mc}
  \log p_{MC}(\mathcal{S}_{i})  =  \sum_{j = 1}^{L}\sum_{m = 1}^{M} \mathbbm{1}(e_j=m)   \log p_m^{*}(t_j) 
  = \sum_{j = 1}^{L}\sum_{m = 1}^{M} \mathbbm{1}(e_j=m)   \log \lambda_m^{*}(t_j)
  -\sum_{m = 1}^{M} \left(   \int_{t_1}^{t_L}  \lambda_m^{*}(s) \,ds  \right).
\end{multline}

Here, $\mathbbm{1}()$ is the indicator function.
The log-likelihood can be explained with the help of two pieces of information: Firstly, $\lambda_m^{*}(t)dt$ represents the likelihood of observing an event of type $m$ in the interval $[t_j,t_j+dt)$ conditioned on past events $\mathcal{H}_{t_j}$. Secondly, we can calculate the likelihood of not witnessing any events of type $m$ in the rest of the interval $[t_1,t_L]$ by computing $\exp \left(-\int_{t_1}^{t_L}  \lambda_m^{*}(s) ds\right)$.
%The log-likelihood can be understood using the following two facts: First, the quantity $\lambda_m^{*}(t)dt$ corresponds to the probability of observing an event of type $m$ in the infinitesimal interval $[t_j,t_j+dt)$ conditioned on the past events $\mathcal{H}_{t_j}$. Second, we can compute the probability of not observing any events of type $m$ in the rest of the interval $[t_1,t_L]$ as $\exp \left(-\int_{t_1}^{t_L}  \lambda_m^{*}(s) ds\right)$.

However, in many cases, such as in EHRs, it is common to have co-occurring events (multi-label setting (ML)). To handle this issue, \cite{enguehardNeuralTemporalPoint2020} proposed using a binary cross-entropy function:

\begin{equation*} \label{eq:CIF-ml}
  \log p_{ML}(\mathcal{S}_{i}) = \log p_{MC}(\mathcal{S}_{i}) 
  + \sum_{m = 1}^{M} (1-\mathbbm{1}(e_j=m)) \log \left(1-p_m^{*}(t_j)  \right).
\end{equation*}

% \begin{multline} 
%     -\log p(\mathcal{S}_{i})  =  - \sum_{j = 1}^{L}\sum_{m = 1}^{M} 1(e_j=m)   \log \lambda_m^{*}(t_j) \\   
%     + \sum_{m = 1}^{M} \left(   \int_{0}^{T}  \lambda_m^{*}(s) \,ds  \right)\\
%     - \sum_{j = 1}^{L}\sum_{m = 1}^{M} (1-1(e_j=m)) \log \left(1-\lambda_m^{*}(t_j) \exp \left( - \int_{0}^{T}  \lambda_m^{*}(s) \,ds  \right) \right)
% \end{multline} 

%  It should be noted that the real advantage of point process is modeling non-event likelihoods in the form of integrals. If we neglect the integrals, we would achieve the cross-entropy and binary cross entropy loss in the multi-class and multi-label settings respectively for the prediction of next mark given history of events.

Another approach is the marked case, which assumes that the marks and timestamps are conditionally independent given the history of events ($\mathcal{H}_t$):

\begin{equation*} \label{eq:CIF-marked}
  \log p_{marked}(\mathcal{S}_{i}) =
  \sum_{j = 1}^{L}\sum_{m = 1}^{M} \mathbbm{1}(e_j=m)   \log p^{*}(e_j=m) 
  +\sum_{j = 1}^{L}\lambda^{*}(t_j) - \int_{t_1}^{t_{L}}\lambda^{*}(t^{\prime})dt^{\prime}.
\end{equation*}

This marked case is essentially an AutoEncoder (AE) for the next mark prediction, with a single-dimensional point process for timestamps only.

It is important to note that the main advantage of point processes lies in their ability to model non-event likelihoods in the form of integrals. If we neglect the integrals in equations (\ref{eq:CIF-mc}, \ref{eq:CIF-ml}, \ref{eq:CIF-marked}), we would end up with the cross-entropy and binary cross-entropy loss in the multi-class and multi-label settings, respectively, for predicting the next mark given the history of events.

\subsubsection*{Neural Point Process}

Encoder-decoder architectures have proven to be effective in many time series applications. The main idea of a neural point process (NPP) is to first encode the history of events until $t_j$ using a neural network architecture as $h_j=Enc(\mathcal{H}_{j+1};\theta)$, and later use $h_j$ to parameterize the CIFs using a different decoder architecture $\lambda^*_m(t)=Dec(h_j;\phi)$ for $ t \in (t_j,t_{j+1}]$ \cite{shchurNeuralTemporalPoint2021}.

% Initial works have used recurrent encoders such as RNN, GRU or LSTM [refs] in which the hidden state gets updated after the arrival of a new event as $h_{j+1}=Update(h_j,(t_j,e_j))$. The main advantage is that they allow us to compute history embeddings in $O(L)$ time, however, they are prone to neglect long-term inter-event influences. On the other hand, set aggregation encoders directly encode all past events into a history embedding. Adopting an attention mechanism is one such way that can capture long-term influences and can be trained in parallel as well which makes it more computationally efficient. For example, \cite{zuoTransformerHawkesProcess2020} proposed Transformer Hawkes Process (THP) that adopts a transformer architecture for event encoding.

Initial works have utilized recurrent encoders such as RNN \cite{duRecurrentMarkedTemporal2016}, or LSTM \cite{meiNeuralHawkesProcess2017}, where the hidden state is updated after the arrival of a new event as $h_{j+1}=Update(h_j,(t_j,e_j))$. The main advantage of these models is their ability to compute history embeddings in $O(L)$ time. However, they are susceptible to ignoring long-term inter-event dependencies. In contrast, set aggregation encoders encode all past events directly into a history embedding. One approach to capture long-term dependencies is to use self-attention \cite{zuoTransformerHawkesProcess2020,zhangSelfAttentiveHawkesProcess2020}, which can be trained in parallel and is more computationally efficient.

We focus on neural point process models based on attention mechanisms that can potentially address the problems of slow serial computing and loss of long-term information. In addition, attention weights bring interpretability and can show peer influences between events. Self-attentive Hawkes Process (SAHP) \cite{zhangSelfAttentiveHawkesProcess2020} proposes a multi-head attention network as the history encoder. In addition, they use a \emph{sofplus} function that can capture both excitation and inhibition effects. Similarly, Transformer Hawkes Process (THP) \cite{zuoTransformerHawkesProcess2020} adopts the transformer architecture \cite{vaswaniAttentionAllYou2017} by using time encodings instead of positional encodings in the original architecture of transformers \cite{vaswaniAttentionAllYou2017}. However, their model only captures mutual excitations between events. In another interesting study \cite{enguehardNeuralTemporalPoint2020}, researchers studied different combinations of encoders (self-attention and GRU) and decoders on various datasets for comparison. They demonstrated that attention-based NPPs appear to transmit pertinent EHR information and perform favorably compared to existing models. One gap in the current literature is that they do not have any means for encoding additional information that can be useful for the characterization of CIFs. For example, in EHRs, there are other sources of information such time series values that could be useful for the characterization of CIFs.

\subsection{Deep learning for irregularly sampled data}

Recurrent neural networks have been modified to consider irregularly sampled time series. For example, GRU-D \cite{cheRecurrentNeuralNetworks2018} adapts GRU to consider missingness patterns in the form of feature masks and time intervals to achieve better prediction results. RETAIN \cite{choiRETAINInterpretablePredictive2017} is based on a two-level neural attention model that is specifically designed for EHRs data. SeFT \cite{hornSetFunctionsTime2020} is based on recent advances in differentiable set function learning \cite{zaheerDeepSets2017}, which is extremely parallelizable with a beneficial memory footprint, thus scales well to large datasets of long time series and online monitoring scenarios. Their use of aggregation function is similar to transformers that compute the embeddings of set elements independently, leading to lower runtime and memory complexity of $O(P)$. Although these models are nearly end-to-end and eliminate the need for an imputation pipeline, it is still unclear how much they are affected by the missingness pattern in EHRs. Additionally, they do not explicitly model the missingness pattern in the data.

% \section{Related Works}
% \label{sec:Related Works}

\section{Proposed Model}

\begin{figure*}[!t]
  \centerline{\includegraphics{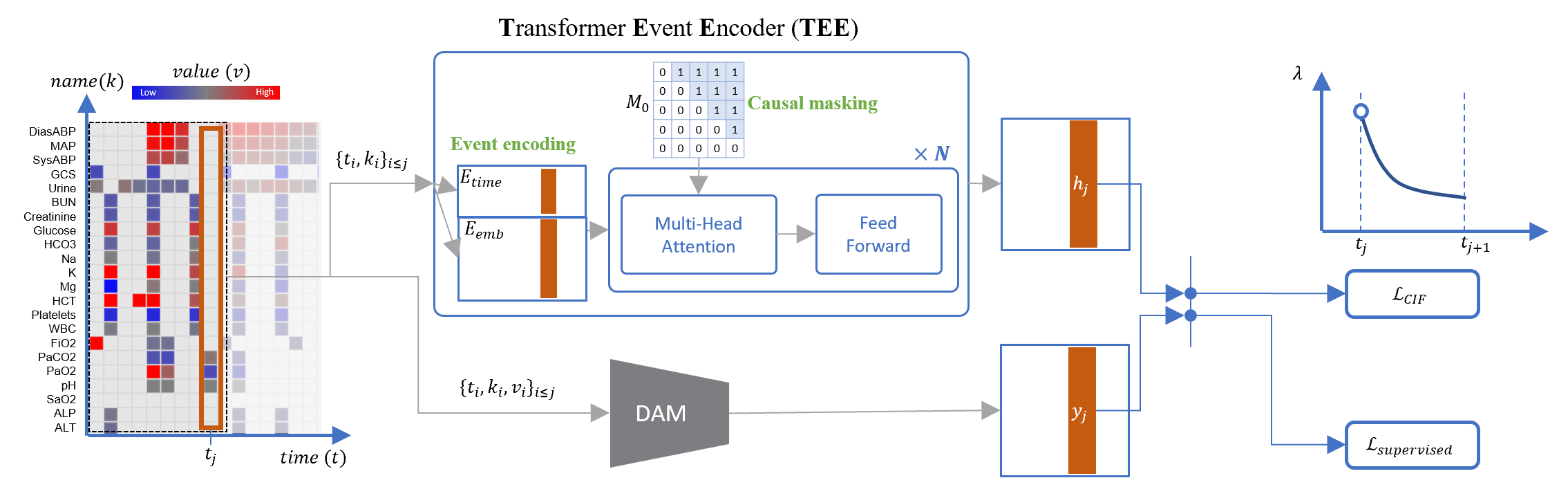}}
  \caption{Schematic representation of \emph{TEE4EHR} model combining a transformer-based event encoder (TEE) with a deep attention module (DAM) that can handle irregularly sampled time series.}
  \label{fig1}
\end{figure*}

The proposed model, \emph{TEE4EHR}, consists of two modules: a Transformer Event Encoder (TEE) for handling event data ($\mathcal{S}_i$) and a Deep Attention Module (DAM) for handling irregularly sampled time series ($\mathcal{U}_i$) \cite{hornSetFunctionsTime2020}. The schematic of the model is depicted in Fig. \ref{fig1}. The irregularly-sampled time series of an example patient is depicted as an image where the x-axis represents time, the y-axis represents the variable and the color indicates the value. Both TEE and DAM modules encode the data from the beginning until $t_j$ into $h_j$ and $y_j$ embeddings respectively. The concatenated vector $[h_j,y_j]$ can be used to parameterize the conditional intensity functions in the next interval or any other downstream tasks.

% encoding a dataset with event sequence data and irregularly sampled time series. 

% The key advantage of our proposed model is to combine a transformer event encoder (TEE) for processing the pattern of laboratory tests with a deep attention module (DAM) \cite{hornSetFunctionsTime2020} that can handle an irregularly sampled time series for any downstream prediction task.

\subsection{Transformer Event Encoder}
\subsubsection*{Architecture}
We use a transformer event encoder (TEE) similar to THP \cite{zuoTransformerHawkesProcess2020} with a few modifications.
In the first step, we embed all event marks as $E_{emb}=E \times W_{emb}$, where $E \in \mathbb{R}^{L \times M}$ is the binary encoding matrix of all event marks (multi-label or multi-class), and  $ W_{emb} \in  \mathbb{R}^{M \times d_{emb} }  $ is the trainable embedding matrix. In addition, we encode the vector of timestamps $t=[t_1,t_2,...,t_L] $ to $E_{time}=[TE(t_1),TE(t_2),...,TE(t_L)]\in \mathbb{R}^{L \times d_{time}}$ using the following transformation formula:

\begin{equation}\label{eq:time-encoding}
  TE(t_j) =
  \begin{cases}
    \cos\left(  \frac{t_j}{\mathcal{T}^{(j-1)/d_{time}} }  \right) & \text{if } j \text{ is odd}  \\
    \sin\left(  \frac{t_j}{\mathcal{T}^{j/d_{time}} }  \right)     & \text{if } j \text{ is even}
  \end{cases}
\end{equation}

% old equatioin. copy above
% [z(t_j)]_{k} =
% \begin{cases}
%     \cos\left(  \frac{t_j}{\mathcal{T}^{(k-1)/d_t} }  \right) & \text{if } k \text{ is odd}  \\
%     \sin\left(  \frac{t_j}{\mathcal{T}^{k/d_t} }  \right)     & \text{if } k \text{ is even}
% \end{cases}

Here, $\mathcal{T}$ represents the maximum time scale and $d_{\text{time}}$ is the time embedding dimension. This transformation closely resembles positional encodings in transformers \cite{vaswaniAttentionAllYou2017}, where the index is replaced by the timestamp.

Unlike THP and the original positional encoding \cite{vaswaniAttentionAllYou2017}, which assume $d_{\text{emb}}=d_{\text{time}}$ and add the time encoding to the event embedding, we propose to concatenate these two vectors before providing them as input to the transformer block (our first modification):

\begin{equation}
  X=[E_{emb}, E_{time}] \in \mathbb{R}^{L \times (d_{emb}+t_{emb})}
\end{equation}

% In the second step, timestamps should be encoded and added to the event embedding, however, we propose to concatenate time encodings that can lead to better characterization of conditional intensity functions. Finally, the input of the transformer encoder will be $X_{emb}=[E_{emb}, T_{emb}] \in \mathbb{R}^{L \times (d_{emb}+t_{emb})}$.

Finally, we use the standard transformer architecture with multiple layers and attention heads to encode the embedded events matrix $X$ into the encoded matrix $ H=(h_1, ..., h_j, ..., h_L) $.

% It is important to use an appropriate mask matrix to prevent information leakage from the future to the past. In this case, $h_j$ should contain all available information until the occurrence of $j$-th event. Later on, we will use $h_j$ to parameterize CIFs in the interval $(t_j,t_{j+1}]$. As a result, the masking matrix $m_0$ should be an upper triangular matrix where the arrays above the diagonal are one. Here, we introduce another variation in which we mask up to a specified number of events. For example, if we perform a column shift by $w$ on $m_0$, we will obtain $m_w$. As a result, $h_j$ would contain the information for the first $(j-w)$ events.

Causal masking is essential in our transformer architecture to prevent information leakage from the future to the past. Specifically, the vector $h_j$ should contain all the available information up to the $j$-th event, which will later be used to parameterize the CIFs within the interval $(t_j,t_{j+1}]$. By default, the masking matrix $M_0$ must be an upper triangular matrix where the elements above the diagonal are all one (one indicates the elements to be masked). Our second modification is to consider an additional masking parameter $w$. For instance, if we perform a left column shift of $w$ on $M_0$, we can obtain $M_w$. This results in $h_j$ that contains the information of the first $(j-w)$ events. This type of masking could prevent overfitting and will be investigated in more detail in our experimental part.

\subsubsection*{Attention Aggregation}

We can use the attention matrix of the transformer event encoder of each sample for model interpretability. In addition, we can aggregate attention matrices of a group of samples to extract an influence matrix that reveals the interaction between events. Consider the attention matrix of $i$-th sample $A^i_{L_i \times L_i}$ where the sum of the elements in each row equals one. we multiply each row by the number of unmasked events to compensate for different event counts. We define the event frequency matrix $C^i$ such that $C^i_{ij}=1$ if $A^i_{jk}>0$. Similarly, the event interaction matrix $I^p$ specifies the significant attention values such that $I^i_{jk}=1$ if $A^p_{jk}>\epsilon$.

Finally, we can aggregate these matrices for N samples:

\begin{gather}
  C^{agg}_{mn} = \left( \sum_{i=1}^{N}  \sum_{k=1}^{L_i}  \sum_{j=1}^{k}  C^p_{jk} \mathbbm{1}(e^i_k=n)\mathbbm{1}(e^i_j=m)   \right)/N \\
  I^{agg}_{mn} = \left( \sum_{i=1}^{N}  \sum_{k=1}^{L_i}  \sum_{j=1}^{k}  I^p_{jk} \mathbbm{1}(e^i_k=n)\mathbbm{1}(e^i_j=m) \right) /(NC^{agg}_{mn})
\end{gather}

Here, $C^{agg}_{mn}$ can be interpreted as the average number of times that the event $n$ occurs before the event $m$. Similarly, $I^{agg}_{mn}$ reveals the fraction of $C^{agg}_{mn}$ in which event $n$ plays a significant role in the prediction of event $m$.

\subsection{Deep Attention Module}

We use a deep attention module (DAM) \cite{hornSetFunctionsTime2020} for encoding all additional information including irregularly sampled time series. Each observation is represented by $u_p=(TE(t_p), k_p, v_p) \in d^s$ where $TE(t_p)$ is the same transformation for time encoding in Equation (\ref{eq:time-encoding}).% we define attention $a(\mathcal{U}_k,s_k )$

We define $\mathcal{U}_p$ to be the set of the first $p$ observations.
The goal is to calculate the attention weight $a(\mathcal{U}_p,u_k ), k \leq p$ that is the relevance of $k$-th observation $u_k$ to $\mathcal{U}_p$.
This is achieved by computing an embedding of the set elements using smaller set functions $f^{\prime}$, and projecting the concatenation of the set representation and the individual set element into a $d$-dimensional space:

\begin{equation}\label{eq:fprime}
  f^{\prime}(\mathcal{U}_p) = g^{\prime} \left(  \frac{1}{|p|} \sum_{u_k \in \mathcal{U}_p}  h^{\prime}(u_k;\theta^{\prime}) ;\rho^{\prime}\right)
\end{equation}

Here, we compute the mean of the first $p$ observations after passing the first $p$ observations through a multilayer perceptron (MLP) neural network ($h^{\prime}(u_k;\theta^{\prime})$). Finally, a second transformation $g^{\prime}$ is performed to obtain embeddings $f^{\prime}(\mathcal{U}_p) \in \mathbb{R}^{d_{g^{\prime}}}$.

Then we can compute the key values ($ K_p$) using the key matrix $W^k \in \mathbb{R}^{(d_{g^{\prime}}+d_s) \times d_{prod}}$:

\begin{equation}
  K_p=[f^{\prime}(\mathcal{U}_p), u_p]^T W^K.
\end{equation}

Using a query vector $w^q \in \mathbb{R}^{d_{prod}}$, we compute the desired attnetion weight:

\begin{equation}
  a(\mathcal{U}_p,u_k)=  softmax(\frac{K_p.w^q}{\sqrt{d_{prod}}  }).
\end{equation}

Finally, we compute a weighted aggregation of set elements using attention weights similar to equation (\ref{eq:fprime}):

\begin{equation*}
  y^{\prime}_p=f(\mathcal{U}_p) =
  g\left(
  \sum_{u_k \in \mathcal{U}_p}  a(\mathcal{U}_p,u_k)h(u_k;\theta);\rho
  \right)
\end{equation*}

We consider $y^{\prime}_p \in \mathbb{R}^{d_g}$ as the representation of the first $p$ observations.

The matrix $Y^{\prime}=[y^{\prime}_1,y^{\prime}_2,...,y^{\prime}_P] \in \mathbb{R}^{P \times d_g}$ keeps all representations of the data. To use state information ($Y^{\prime}$) and event encodings ($X$) for CIFs characterization, we down-sample $Y^{\prime}$ to $Y=[y_1,y_2,...,y_L] \text{ where } y_j=y^{\prime}_{p^{*}},  p^{*}=argmax_{t_p \leq t_j} p$.
% \in \mathbb{R}^{L \times d_g}

Without loss of generality, we can also consider multiple heads by adding another dimension to keys and queries to create a more complex model. We also embed static variables $d_i$ using a separate MLP module and concatenate it with $y_p$.

% We need to combine event embeddings $ H_{L \times d_e}$ and state embeddings $Y_{P \times d_g}$, however, the length of each matrices does not match. As a result, we consider the reduced version of state matrix as below:

% \begin{equation*}
%     y^{\prime}_j=y_p \text{ where } p=\argmax_{t_p \leq t_j} p
% \end{equation*}

\subsection{Event Decoder}
Once we obtain a representation of a sample using embedded events ($h_j$) and states($y_{j}$), we can parameterize conditional intensity functions (CIFs) of the events.

In neural point process literature, many approaches have been proposed to decode either conditional or cumulative intensity functions. We will use a decoder similar to \cite{zhangSelfAttentiveHawkesProcess2020} as it can model both exciting and inhibiting effects for CIFs:

% \begin{gather*} 
%    \mu_{m,i+1}=gelu(h_{i+1}W_{m,\mu}), 
% \end{gather*}

\begin{gather}
  \mu_{m,j}=gelu(h_{j}W_{m,\mu}  +  y_{j}W_{m,\mu}), \\
  \eta_{m,j}=gelu(h_{j}W_{m,\eta}  +  y_{j}W_{m,\eta}), \\
  \gamma_{m,j}=gelu(h_{j}W_{m,\gamma}  +  y_{j}W_{m,\gamma}).
\end{gather}

The function \emph{gelu} represents the Gaussian Error Linear Unit for nonlinear activations which has been empirically proved to be superior to other activation functions for self-attention \cite{hendrycksGaussianErrorLinear2020}. Finally, we can express the intensity function as follows:

\begin{equation*}
  \lambda_m(t)=\text{softplus}(\mu_{m,j}+\\
  (\eta_{m,j}- \mu_{m,j}) \exp(-\gamma_{m,j}(t-t_j))    ),
\end{equation*}

for $t \in (t_j, t_{j+1}]$, where the \emph{softplus} is used to constrain the intensity function to be positive. Here, $\eta_{m,j}$ is the initial intensity at $t_j$, $\mu_{m,j}$ is the baseline intensity when $t \to \infty$, and $\gamma_{m,j}$ controls the decay rate.

\section{Experiments}
\label{sec:Experiments}

We conduct various experiments to empirically demonstrate the effectiveness of each component in our model. First, we investigate the performance of our proposed TEE module compared to baseline models in the neural point process literature on four common event sequence datasets. Second, we show the advantage of our main model, \emph{TEE4EHR}, on two real-world healthcare datasets for handling irregularly sampled time series.

\subsection{Datasets}
The overal characteristics of the datasets are summarized in Table \ref{tab:data}. The event sequence datasets are as follows:

\textbf{ReTweets (RT)} includes sequences of tweets, with each sequence consisting of an original tweet and subsequent follow-up tweets. For each tweet, the time of posting and the user tag is recorded. Additionally, users are classified into three groups ('small', 'medium', and 'large') based on their number of followers. We use two versions of this dataset for multi-class (RT-MC) \cite{zhangSelfAttentiveHawkesProcess2020} and multi-label (RT-ML) \cite{enguehardNeuralTemporalPoint2020} scenario.

\textbf{Stack Overflow (SO)} is an online platform where people can ask questions and get answers. To encourage active participation, users are awarded badges, and it is possible for a user to receive the same badge multiple times. data is collected over the span of two years and each user's badge history is considered as a sequential sequence. Each event in the sequence represents the receipt of a specific medal. We used the same processed dataset in the literature \cite{zuoTransformerHawkesProcess2020}.

\textbf{Synthea(SYN)} is a synthetic patient-level EHR that is generated using human expert-curated Markov processes \cite{walonoskiSyntheaApproachMethod2018}. Here, we reused the already processed version of this data by \cite{enguehardNeuralTemporalPoint2020}.

We have also used two real-world ICU datasets from PhysioNet \cite{PhysioBankPhysioToolkitPhysioNet}:

\textbf{Physionet 2012 Mortality Prediction Challenge (\emph{P12})} contains 12,000 ICU stays each of which lasts at least 48 hours  \cite{silvaPredictingInHospitalMortality}. At admission, general descriptors like gender or age are recorded for each stay. Throughout the stay, up to 37 time-series variables, such as blood pressure, Lactate, and respiration rate, are measured depending on the patient's condition. These variables may be measured at regular intervals, like hourly or daily, or only when necessary. The task is to predict in-hospital moratility and the prevalence is 14.2\%.

\textbf{Physionet 2019 Sepsis Early Prediction Challenge (\emph{P19})} includes approximately 40,000 patients in the ICU \cite{reynaEarlyPredictionSepsis2020}. The data format is similar to P12 and the task is to predict whether a patient will develop sepsis within 6 hours. The prevalence of patients with sepsis shock is 4.2\%.

% Table generated by Excel2LaTeX from sheet 'datasets'
\begin{table*}[htbp]
  \centering
  \caption{Datasets description}

  % Table generated by Excel2LaTeX from sheet 'datasets'
\begin{tabular}{ccccc}
\toprule
\toprule
Dataset & Task  & \#events & \#samples & Avg. length \\
\midrule
Stack Overflow (SO) & Multi-class & 22    & 6.6k  & 72 \\
ReTweets (RT-ML) & Multi-class & 3     & 24k   & 109 \\
ReTweets (RT-MC) & Multi-label & 3     & 20k   & 104 \\
Synthea (SYN) & Multi-label & 357   & 12k   & 43 \\
Physionet 2012 (P12) & Multi-label & 24    & 12k   & 20 \\
Physionet 2019 (P19) & Multi-label & 25    & 38k   & 8.3 \\
\bottomrule
\bottomrule
\end{tabular}%

  \label{tab:data}%

\end{table*}%

% \subsection{Setups}

\subsection{Preliminary evaluation of TEE}

In the first series of experiments on event sequence datasets, we investigate TEE module with three different loss functions:

\begin{itemize}

  \item \emph{PP(multi-class or multi-label)} uses multi-class or multi-label point process loss (equations (\ref{eq:CIF-mc}) and (\ref{eq:CIF-ml}) respectively).

  \item \emph{PP(marked)} is a marked point process (\ref{eq:CIF-marked}) where we assume marks and time stamps are independent.

  \item \emph{AE} is a simple auto-encoder for predicting future events from event embeddings without CIF characterization. The loss function is cross-entropy or binary cross-entropy for multi-class and multi-label datasets respectively. We use this loss function as an ablation study to fully understand the utility of integral terms in the point process loss functions.
\end{itemize}

Throughout the experiments, time concatenation and a masking parameter of $w=1$ are found to be more effective than other values which are used as the default setting for all of our experiments. However, we demonstrate the performance gain with different time-encoding strategies (concatenation and summation) and different values for the masking parameter ($w \in [0,1,2,3]$) to see the independent effect of each component.

\subsection{TEE for EHRs}

After evaluating our TEE module using event sequence datasets, we aim to demonstrate its effectiveness in two real-world healthcare datasets, P12 and P19. Initially, we utilize both TEE and DAM to model the patterns of laboratory tests, which serve as our events, by characterizing their conditional intensity functions. Notably, this approach operates as a form of self-supervised learning, as we do not explicitly rely on any labels. In the subsequent phase, we employ the transferred TEE module from the initial step alongside DAM for outcome prediction, transitioning into a supervised learning framework. Our objective is to assess the potential utility of the TEE module, which encodes the patterns of laboratory tests, for accurate outcome prediction.

\subsubsection*{Self-supervised Learning}

In this step, we train TEE together with DAM using different loss functions (equations \ref{eq:CIF-mc}-\ref{eq:CIF-marked}). Here, we have two main differences compared to the neural point process framework. First, we use DAM separately for embedding irregularly sampled time series. Second, we use the embeddings of both TEE and DAM to parameterize CIFs. To assess the effectiveness of jointly learning DAM, we compare it with the case where we train TEE alone.

Here, we regard the occurrence of certain laboratory variables (or patterns of irregularly sampled laboratory tests) as events that are part of routine care (for more details, refer to Appendix A).

To assess the quality of learned representations during the self-supervised learning step, we use a simple multilayer perceptron layer for predicting the outcome from embeddings. This layer is detached from the main network and does not affect the training of TEE and DAM.

% We also train an additional label prediction module based on the learned representations. This module is a simple MLP detached from TEE and DAM, meaning that the label loss does not affect either of the TEE or DAM modules. We do this to assess the predictive capacity of the learned representations during the self-supervised learning step.

\subsubsection*{Supervised Learning}
In the last series of experiments, we investigate the utility of learned TEE in the self-supervised learning step in a downstream task for predicting sepsis shock and hospital mortality in P19 and P12 respectively. In particular, we transfer and freeze the learned TEE so that it wouldn't be affected by the supervised loss function.

% Here again, we consider different loss functions as explained in the first series of experiments. As an ablation study, we report the results for DAM only as well as the cases where we train TEE and DAM jointly from scratch. Training details can be found in Appendix B.

\subsection{Baselines}

For the preliminary experiments, we compare TEE with the following baselines: SAHP \cite{zhangSelfAttentiveHawkesProcess2020}, which was the first model to utilize attention weights; THP \cite{zuoTransformerHawkesProcess2020}, which was the first work to introduce transformers to event sequence data; Latent graphs \cite{zhangLearningNeuralPoint2021}, which is based on a probabilistic graph generator; and GRU-CP \cite{enguehardNeuralTemporalPoint2020} which uses a GRU decoder with conditional Poisson (CP) decoder for CIF characterization.

For the supervised learning tasks, we employ state-of-the-art deep learning models that handle irregularly sampled time series: Transformer \cite{vaswaniAttentionAllYou2017}, GRU-D \cite{cheRecurrentNeuralNetworks2018}, SeFT \cite{hornSetFunctionsTime2020}, IP-NET \cite{shuklaInterpolationPredictionNetworksIrregularly2019}, mTAND \cite{shuklaMultiTimeAttentionNetworks2020}, and RainDrop \cite{zhangGraphGuidedNetworkIrregularly2022}. We use the implemented version of these models by RaindDrop\footnote{\url{https://github.com/mims-harvard/Raindrop/}}.

\subsection{Metrics}
We report log-likelihood normalized by the number of events (LL/\#events) as a goodness of fit for CIFs characterization \cite{zhangSelfAttentiveHawkesProcess2020,zuoTransformerHawkesProcess2020}. However, we do not compare this metric in the preliminary evaluation of TEE as it is problematic to compare across different loss functions (equations \ref{eq:CIF-mc}-\ref{eq:CIF-marked}) and models with different event decoders.

For future event type prediction, we report the weighted measure of F1-score and area under the receiver operating characteristic curve (AUROC) in the multi-class and multi-label setting, respectively. In the supervised learning task for binary prediction, we report F1-score and area under the precision-recall curve (AUPRC), and AUROC.

We use t-Distributed Stochastic Neighbor Embedding (t-SNE) to show learned representations in the downstream tasks, which is a machine learning algorithm used to visualize high-dimensional data in a lower-dimensional space \cite{maatenVisualizingDataUsing2008}.

We also introduce a similarity metric between the pattern of laboratory tests of a patient and its 10 nearest neighbors in the embedding space (learned representations). To compute this metric, we first calculate the \emph{measurement density} of each laboratory variable during the patient stay:

\begin{equation}
  d_{m}^{i}=  \frac{1}{t_L} \sum_{j=1}^{L} \mathbbm{1}(e_j=m).
\end{equation}

Here, $d_{m}^{i}$ represents the average frequency of the $m$-th laboratory variable in one hour for the $i$-th patient. Then, for each patient $i$, we calculate the cosine similarity between $d^i$ and each of its 10 closest neighbors in the embedding space and determine its average ($CS^{i}_{avg}$).
The average of $CS^{i}_{avg}$ for all positive patients is referred to as the 10 nearest neighbor pattern similarity (10nn-ps) and is reported for each dataset.

% \begin{equation}
%     CS_{avg}^{i}=  \frac{1}{10} \sum_{j=1}^{10} \frac{d^i.d^j}{|d^i||d^j|}.
% \end{equation}

\section{Results \& Discussion}

\subsection{Preliminary evaluation of TEE}

% Table generated by Excel2LaTeX from sheet 'T1_new'
\begin{table*}[htbp]
  \centering
  \caption{Performance of TEE on common event sequence datasets. For each metric, mean(std) across splits are reported. The best results are highlighted in bold. NA: not applicable. NR: not reported in the original work.}

  % Table generated by Excel2LaTeX from sheet 'PRE'
\begin{tabular}{ccccccc}
\toprule
\toprule
      &       &       & \multicolumn{4}{c}{\textbf{Dataset}} \\
\cmidrule{4-7}      &       &       & \textbf{RT-MC} & \textbf{RT-ML} & \textbf{SO} & \textbf{SYN} \\
\multicolumn{3}{c}{\textbf{Model}} & \textbf{F1-Score} & \textbf{AUROC} & \textbf{F1-Score} & \textbf{AUROC} \\
\midrule
\multicolumn{3}{c}{Latent} & 0.58  & NR    & 0.28(0.19) & NR \\
\multicolumn{3}{c}{SAHP} & 0.54  & NR    & 0.24  & NR \\
\multicolumn{3}{c}{THP} & 0.54  & NR    & 0.24  & NR \\
\multicolumn{3}{c}{GRU-CP} & NR    & 0.61  & 0.33  & 0.85 \\
\multicolumn{3}{c}{TEE with AE loss} & \textbf{0.63} & 0.70(0.03) & \textbf{0.38(0.02)} & \textbf{0.90(0.01)} \\
\multicolumn{3}{c}{TEE with PP(ML/MC)} & 0.53  & \textbf{0.74(0.02)} & 0.32(0.00) & 0.60(0.05) \\
\multicolumn{3}{c}{TEE with PP(single)} & 0.57  & 0.70(0.03) & 0.36(0.01) & 0.64(0.04) \\
\bottomrule
\bottomrule
\end{tabular}%

  \label{tab:pre}%
\end{table*}%

Table \ref{tab:pre} shows the performance of our proposed TEE with different loss functions compared to various baselines. We can see that TEE with AE loss achieves an F1-score of 0.63 and 0.38 for RT-MC and SO respectively, and an AUROC of 0.90 for the SYN dataset. TEE with PP(ML) loss function achieves an AUROC of 0.74 for RT-ML. All of the achieved scores show improvements compared to the baselines.

The Stack Overflow and Retweets datasets have been widely used in the literature on neural point processes to demonstrate the advantages of modeling non-event likelihoods using the point process loss function. However, we show for the first time that a simple autoencoder model can perform even better. RT-ML is the only dataset where the PP(ML) loss function outperforms the AE loss. Upon examining this dataset, we observe that the occurrence of the same event in consecutive sequences (self-exciting behavior) is more frequent, which may explain why the point process loss function achieves better results than the AE loss. As we mentioned previously, the point process loss is the sum of the AE loss and an integral term that models non-event likelihoods. Therefore, for future studies on neural point processes, we recommend researchers compare their proposed models against ablated versions where the integral term is omitted.

\begin{table*}[htbp]
  \centering
  \caption{Performance of TEE on common event sequence datasets. For each metric, mean(std) across splits are reported. The best results are highlighted in bold.}

  % Table generated by Excel2LaTeX from sheet 'Sheet11'
\begin{tabular}{cccccccc}
\toprule
\toprule
      &       &       &       & \multicolumn{4}{c}{\textbf{Dataset}} \\
\cmidrule{5-8}      &       &       &       & \textbf{RT-MC} & \textbf{RT-ML} & \textbf{SO} & \textbf{SYN} \\
\multicolumn{3}{c}{\textbf{TEE}} &       & \textbf{F1-Score} & \textbf{AUROC} & \textbf{F1-Score} & \textbf{AUROC} \\
\cmidrule{1-3}\cmidrule{5-8}\textbf{Loss} & \textbf{w} & \textbf{time\_enc} &       &       &       &       &  \\
\cmidrule{1-3}\multirow{8}[8]{*}{AE} & \multirow{2}[2]{*}{3} & concat &       & \textbf{0.63} & 0.71(0.02) & \textbf{0.36(0.01)} & \textbf{0.90(0.00)} \\
      &       & sum   &       & 0.62  & 0.69(0.03) & 0.36(0.01) & 0.89(0.01) \\
\cmidrule{2-8}      & \multirow{2}[2]{*}{2} & concat &       & \textbf{0.63} & 0.71(0.02) & \textbf{0.37(0.01)} & \textbf{0.90(0.01)} \\
      &       & sum   &       & 0.62  & 0.69(0.02) & 0.35(0.02) & 0.89(0.01) \\
\cmidrule{2-8}      & \multirow{2}[2]{*}{1} & concat &       & \textbf{0.63} & 0.70(0.03) & \textbf{0.38(0.02)} & \textbf{0.90(0.01)} \\
      &       & sum   &       & 0.61  & 0.69(0.03) & 0.34(0.01) & 0.86(0.06) \\
\cmidrule{2-8}      & \multirow{2}[2]{*}{0} & concat &       & 0.59  & 0.66(0.03) & 0.32(0.01) & 0.84(0.05) \\
      &       & sum   &       & 0.59  & 0.66(0.03) & 0.33(0.00) & 0.87(0.03) \\
\midrule
\multirow{8}[8]{*}{PP(ML/MC)} & \multirow{2}[2]{*}{3} & concat &       & 0.53  & \textbf{0.74(0.01)} & 0.33(0.01) & 0.58(0.02) \\
      &       & sum   &       & 0.35  & 0.72(0.01) & 0.33(0.01) & 0.59(0.05) \\
\cmidrule{2-8}      & \multirow{2}[2]{*}{2} & concat &       & 0.52  & \textbf{0.74(0.01)} & 0.33(0.01) & 0.60(0.02) \\
      &       & sum   &       & 0.35  & 0.72(0.01) & 0.32(0.00) & 0.61(0.04) \\
\cmidrule{2-8}      & \multirow{2}[2]{*}{1} & concat &       & 0.53  & \textbf{0.74(0.02)} & 0.32(0.00) & 0.60(0.05) \\
      &       & sum   &       & 0.45  & 0.71(0.02) & 0.32(0.00) & 0.60(0.05) \\
\cmidrule{2-8}      & \multirow{2}[2]{*}{0} & concat &       & 0.44  & 0.66(0.03) & 0.32(0.01) & 0.57(0.06) \\
      &       & sum   &       & 0.44  & 0.66(0.03) & 0.32(0.00) & 0.61(0.05) \\
\midrule
\multirow{8}[8]{*}{PP(single)} & \multirow{2}[2]{*}{3} & concat &       & 0.57  & 0.71(0.02) & 0.37(0.01) & 0.65(0.04) \\
      &       & sum   &       & 0.56  & 0.68(0.04) & 0.35(0.00) & 0.65(0.05) \\
\cmidrule{2-8}      & \multirow{2}[2]{*}{2} & concat &       & 0.53  & 0.70(0.03) & 0.38(0.01) & 0.65(0.05) \\
      &       & sum   &       & 0.54  & 0.68(0.03) & 0.35(0.01) & 0.64(0.06) \\
\cmidrule{2-8}      & \multirow{2}[2]{*}{1} & concat &       & 0.57  & 0.70(0.03) & 0.36(0.01) & 0.64(0.04) \\
      &       & sum   &       & 0.54  & 0.68(0.03) & 0.34(0.00) & 0.64(0.04) \\
\cmidrule{2-8}      & \multirow{2}[2]{*}{0} & concat &       & 0.56  & 0.66(0.03) & 0.33(0.00) & 0.62(0.05) \\
      &       & sum   &       & 0.32  & 0.66(0.03) & 0.33(0.00) & 0.62(0.04) \\
\bottomrule
\bottomrule
\end{tabular}%

  \label{tab:pre-effects}%
\end{table*}%

Table \ref{tab:pre-effects} presents the effect of different time-encoding strategies and masking parameters on the performance of TEE with different loss functions. In general, time concatenation and a masking parameter of $w=1$ yield the best results.

While adding time encodings to event embeddings has been the common practice in the point process literature \cite{zhangSelfAttentiveHawkesProcess2020, zuoTransformerHawkesProcess2020}, we demonstrate for the first time that using time concatenation can lead to improved results in terms of log-likelihood and prediction of the next event type. In natural language processing tasks, summation is typically preferred over concatenation due to its lower memory requirements, fewer parameters, and reduced runtime.
With concatenation, the model already has access to the positional encodings, whereas, with the summation, the model must disentangle the positional information within the hidden layers \cite{wangWhatPositionEmbeddings2020}. We speculate that the datasets used in our study may not be as comprehensive as NLP datasets, hence why concatenation might potentially yield better performance.
We recommend researchers experimenting with the time concatenation strategy on their datasets to determine if it indeed enhances the results.

By using additional masking in our TEE, we effectively regularize the model by limiting its access to future information during training. This constraint can help prevent the model from memorizing specific patterns or sequences in the training data that may not generalize well to future events. Consequently, it encourages the model to learn more robust and generalized representations.
This is very similar to time series forecasting, where additional masking has been shown to improve the performance of the model \cite{zerveasTransformerbasedFrameworkMultivariate2021,wangMultivariateTimeSeries2022}.
We recommend that future research consider the masking parameter as a hyperparameter to be optimized.

\subsection{Self-supervised learning}

% Table generated by Excel2LaTeX from sheet 'T2_new'
\begin{table*}[htbp]
  \centering
  \caption{Self-supervised learning. For each metric, mean(std) across splits are reported. The best results are highlighted in bold. NA: not applicable.}

  % Table generated by Excel2LaTeX from sheet 'T2_new'
\begin{tabular}{ccp{6.84em}p{6.5em}p{7.895em}p{7.945em}}
\toprule
\toprule
      &       & \multicolumn{1}{c}{} & \multicolumn{3}{p{22.34em}}{\textbf{Metric}} \\
\cmidrule{4-6}\multicolumn{1}{p{4.13em}}{\textbf{Dataset}} & \multicolumn{1}{p{4.13em}}{\textbf{Loss}} & \textbf{Model} & \textbf{LL/\#events\newline{}(Future Event)} & \textbf{AUROC\newline{}(Future Event)} & \textbf{AUPRC\newline{}(Label)} \\
\midrule
\multicolumn{1}{c}{\multirow{6}[6]{*}{P12}} & \multicolumn{1}{c}{\multirow{2}[2]{*}{PP(ML)}} & TEE   & -2.51(0.03) & 78.79(0.71) & 27.68(2.53) \\
      &       & TEE+DAM & -2.48(0.41) & 74.69(13.80) & 25.13(6.17) \\
\cmidrule{2-6}      & \multicolumn{1}{c}{\multirow{2}[2]{*}{AE}} & TEE   & NA    & 81.62(0.32) & 28.72(2.82) \\
      &       & TEE+DAM & NA    & 84.61(0.70) & 28.59(1.48) \\
\cmidrule{2-6}      & \multicolumn{1}{c}{\multirow{2}[2]{*}{PP(single)}} & TEE   & -1.35(0.02) & 82.34(0.26) & 30.49(2.04) \\
      &       & TEE+DAM & \textbf{-1.33(0.02)} & \textbf{84.84(0.77)} & \textbf{30.93(2.93)} \\
\midrule
\multicolumn{1}{c}{\multirow{6}[6]{*}{P19}} & \multicolumn{1}{c}{\multirow{2}[2]{*}{PP(ML)}} & TEE   & -1.72(0.09) & 76.81(1.06) & 12.21(1.86) \\
      &       & TEE+DAM & -1.42(0.12) & 77.75(1.61) & 24.19(7.57) \\
\cmidrule{2-6}      & \multicolumn{1}{c}{\multirow{2}[2]{*}{AE}} & TEE   & NA    & 90.42(0.33) & 17.79(4.13) \\
      &       & TEE+DAM & NA    & \textbf{92.30(0.08)} & 26.90(3.12) \\
\cmidrule{2-6}      & \multicolumn{1}{c}{\multirow{2}[2]{*}{PP(single)}} & TEE   & -0.71(0.05) & 90.31(0.32) & 19.41(2.65) \\
      &       & TEE+DAM & \textbf{-0.58(0.03)} & 91.11(0.54) & \textbf{28.99(4.07)} \\
\bottomrule
\bottomrule
\end{tabular}%

  \label{tab:state}%
\end{table*}%

\begin{figure*}[!ht]
  \centering
  % \includesvg[width=0.6\columnwidth]{images/agg2.pdf}
  % \centerline{\includegraphics{images/agg2.png}}

  \centerline{\includegraphics[width=0.95\textwidth]{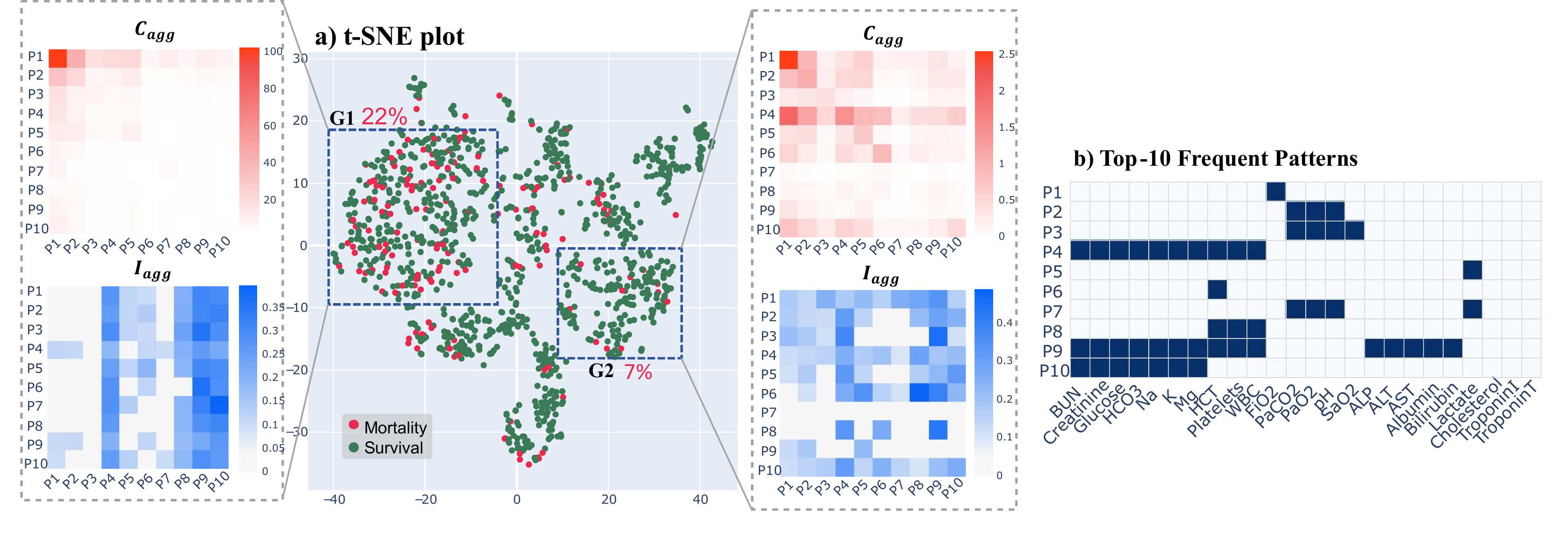}}

  % \includesvg[width=0.6\columnwidth](images/agg2.svg)
  \caption{Attention aggregation in P12 dataset. (a) t-SNE plot of patients with the aggregated frequency matrix and aggregated interaction matrix for two subgroups of patients. (b) Top ten frequent measurement patterns for laboratory tests.}
  \label{fig:agg}

\end{figure*}

\subsubsection{Future pattern prediction}

The results for the prediction of future laboratory tests are reported in Table \ref{tab:state} for different loss functions as well as different architectures (TEE and TEE+DAM). The full model (TEE+DAM) with PP(signle) and AE loss achieves the best AUROC in P12 and P19 respectively.

In all cases, adding DAM results in higher AUROC and LL/\#events for CIF characterization. In the hospital setting, it is plausible that the absolute value of patient states could be advantageous for predicting the order of future laboratory events. To the best of our knowledge, this is the first work that investigates the effectiveness of a sequential neural network (DAM in our work) for characterizing CIFs as well as future event prediction. Although we have evaluated our model using a healthcare database, we believe that a further detailed assessment could be performed on other event sequence data that includes additional information.

We have also reported the AUPRC of the detached label prediction layer from the learned representation. We can see that the TEE+DAM model with PP(single) loss has the highest AUPRC. This indicates that although our model is not trained for the labels, the learned representations from the self-supervised learning approach have predictive value for the patient outcome in the ICU.

% Table generated by Excel2LaTeX from sheet 'Sheet3'
\begin{table*}[!ht]
  \centering
  \caption{Supervised Learning results. For each metric, mean(std) across splits are reported. The best and second-best results are highlighted in bold and underlined respectively. [MODEL] indicates the transferred module from the self-supervised learning task.}

  %   \import{./tables}{downstream.tex}
  % \import{./tables}{T3_june23.tex}

  % Table generated by Excel2LaTeX from sheet 'T3_new'
\begin{tabular}{clccccccc}
\toprule
\toprule
      &       & \multicolumn{3}{c}{P12} &       & \multicolumn{3}{c}{P19} \\
\cmidrule{3-5}\cmidrule{7-9}      &       & \textbf{AUROC} & \textbf{AUPRC} & \textbf{F1-Score} &       & \textbf{AUROC} & \textbf{AUPRC} & \textbf{F1-Score} \\
\midrule
\multirow{6}[2]{*}{\begin{sideways}Baselines\end{sideways}} & Transformer & 84.1(0.9) & 48.5(1.3) & 46.3(3.6) &       & 80.6(2.8) & 43.0(3.8) & 42.7(4.4) \\
      & GRU-D & 81.9(2.1) & 46.1(4.7) & 45.1(2.2) &       & 83.9(1.7) & 46.9(2.1) & 43.9(2.1) \\
      & SeFT  & 74.4(1.3) & 30.2(2.5) & 36.7(1.1) &       & 80.9(2.6) & 41.7(3.4) & 25.2(2.8) \\
      & mTAND & 84.2(0.8) & 48.2(3.4) & 45.9(1.8) &       & 84.4(1.3) & 50.6(2.0) & 45.0(2.9) \\
      & IP-Net & 82.6(1.4) & 47.6(3.1) & 46.0(1.9) &       & 84.6(1.3) & 38.1(3.7) & 41.1(2.3) \\
      & Raindrop & 82.8(1.7) & 44.0(3.0) & 45.0(1.3) &       & 87.0(2.3) & \textbf{51.8(5.5)} & 47.3(3.1) \\
\midrule
\multirow{8}[2]{*}{\begin{sideways}Ours\end{sideways}} & DAM   & 85.7(0.6) & \uline{53.6(1.9)} & 49.8(1.0) &       & 87.6(2.5) & 48.6(5.5) & 47.7(3.1) \\
      & TEE+DAM & 84.4(1.5) & 50.9(4.0) & 48.6(1.8) &       & 87.5(2.2) & 49.6(5.5) & 48.4(4.5) \\
      & TEE+DAM w AE & \textbf{86.2(0.8)} & \uline{53.6(4.2)} & \textbf{51.8(0.6)} &       & \textbf{88.8(2.2)} & 50.3(4.6) & 49.2(3.9) \\
      & TEE+DAM w PP(ML) & 84.9(1.2) & 49.5(5.3) & 48.6(2.3) &       & 87.6(1.8) & 50.0(4.7) & \uline{49.9(1.6)} \\
      & TEE+DAM w PP(Single) & \uline{85.8(0.8)} & \textbf{53.7(2.5)} & 50.9(1.5) &       & \uline{88.7(1.7)} & 50.4(5.1) & 49.4(3.8) \\
      & [TEE w AE]+DAM & \uline{85.8(1.0)} & 52.9(3.0) & \uline{51.6(0.8)} &       & 88.1(2.6) & 48.9(5.8) & 48.1(4.2) \\
      & [TEE w PP(ML)]+DAM & 85.1(0.4) & 50.3(4.1) & 48.8(1.1) &       & 87.6(2.1) & 49.5(3.6) & 48.7(1.8) \\
      & [TEE w PP(Single)]+DAM & 85.4(1.5) & 52.4(3.5) & 50.6(1.7) &       & 88.1(1.9) & \uline{50.6(4.3)} & \textbf{50.4(2.3)} \\
\bottomrule
\bottomrule
\end{tabular}%

  \label{tab:downstream}%
\end{table*}%

\subsubsection{Model interpretability}

Fig.\ref{fig:agg} shows the results for our attention aggregation algorithm in P12 dataset. Fig.\ref{fig:agg}-a illustrates the t-SNE plot for the learned representations of all patients in the self-supervised learning task, where each data point is colored by its true label (red points are positive patients with mortality). Two subgroups of patients are selected for the aggregation analysis. The first subgroup (G1) contains more positive patients (22\%) compared to the second subgroup (G2) with 7\%. This indicates that the self-supervised learning of sampling patterns can distinguish different classes in our dataset.

Furthermore, We select the 10 most frequent patterns of laboratory tests as events (Fig.\ref{fig:agg}-b). For example, the second most frequent pattern (P2) consists of (PaCO2, PaO2, pH) laboratory tests. We show the matrices of aggregated event frequency ($C_{agg}$) and aggregated event interaction ($I_{agg}$) for G1 and G2. In G1, $C_{agg}$ is dominated by FiO2-FiO2 which is expected since the patients in this group are more likely to be intubated and have longer stay in the ICU. In G2, however, $C_{agg}$ reveals other event frequencies with lower magnitude compared to G1. $I_{agg}$ reveals totally different insights. In G1, patterns with more measurements (P4-P8-P9-P10) influence the other patterns with fewer measurements. In G2, however, the influences are smaller and more scattered. In this work, we considered patterns of laboratory tests as events, and explaining the event interaction with our knowledge might not be intuitive. However, our TEE can encode other medical events such as medications, procedures, and diagnoses. In this case, the aggregated event interaction matrix can reveal more interesting insights from a clinical perspective.

\subsection{Supervised Learning}
\subsubsection{Model performance}

The results of outcome prediction (mortality and sepsis shock for P12 and P19 respectively) are reported in Table \ref{tab:downstream}. TEE module indicated inside brackets is the transferred module from the self-supervised learning task. For P12 dataset, one of our model's variations consistently performs better than baselines in all metrics. Within different variations of our proposed models, TEE+DAM with AE and PP(single) losses that are trained from scratch have better performance than the models with transferred TEE from the self-supervised learning step as well as DAM-only model. In P19, the RaindDrop model performs the best in AUPRC only while two variations of our model (TEE+DAM with AE and [TEE w PP(Single)]+DAM) perform better in AUROC and F1-score.

In our experiments, the three transferred modules do not consistently improve over the non-transferred models. One possible reason could be the small size of the data in the pretraining task, especially in the P12 dataset with approximately 12,000 samples. In P19, however, the size of the data is larger (around 38,000 samples) and we observe that the transferred modules enhance performance in AUPRC and F1-Score compared to models trained from scratch. One advantage of transfer learning in this application is that we ensure that the TEE module is not biased by the labels, which could be problematic in healthcare applications \cite{ghassemiReviewChallengesOpportunities2020,haneuseLearningMissingData2016}.

\subsubsection{Learned Representions}

Despite minor improvements in outcome prediction, we contend that the utilization of TEE leads to more effective representation learning in our EHRs. Table \ref{tab:knn} shows the \emph{10nn-ps} metric for different variations of our models in P12 and P19 datasets. We can see that adding TEE module with AE or PP(single) losses can significantly increase our similarity metric compared to the DAM-only model.

\begin{figure}[!h]

  \centerline{\includegraphics{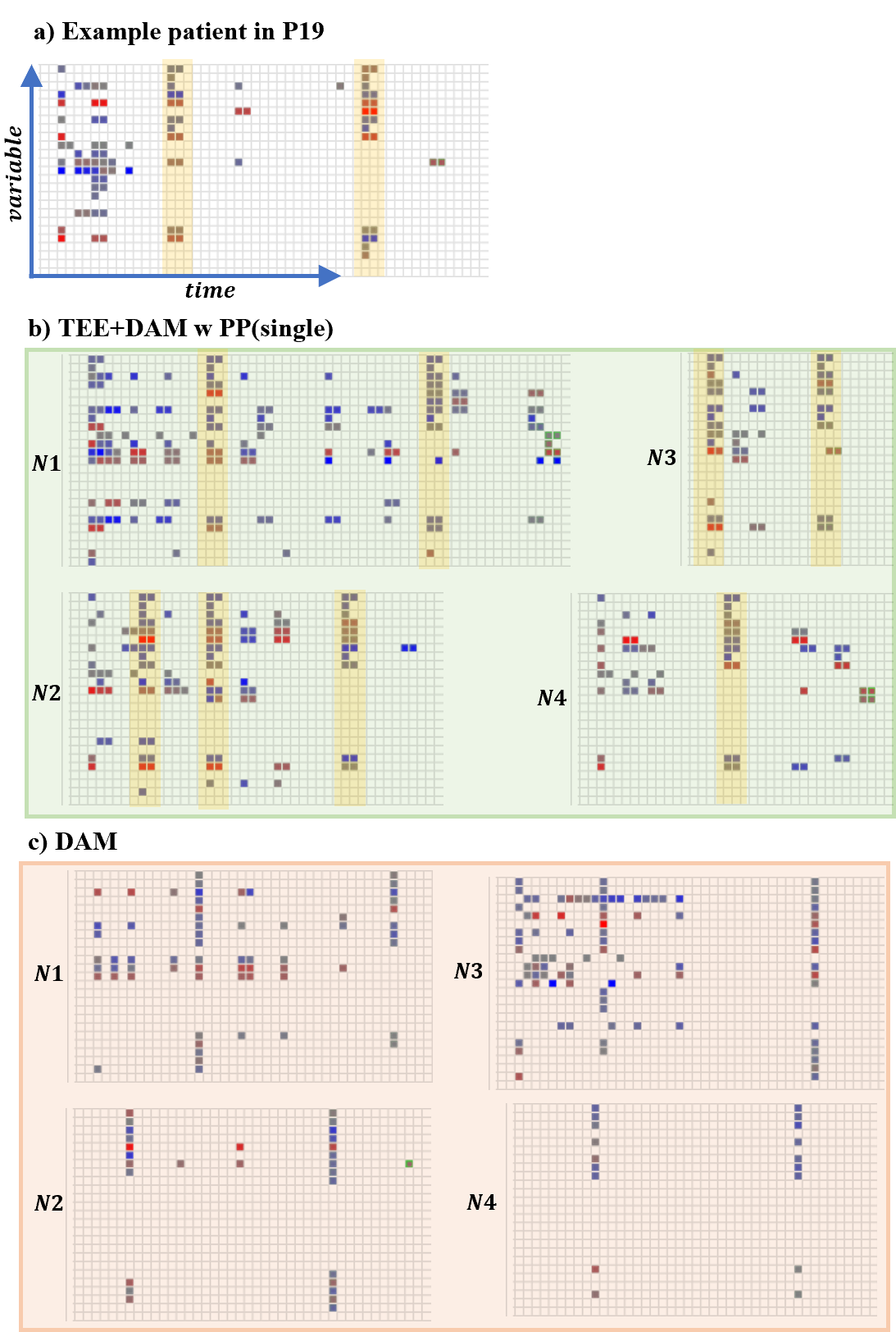}}
  \caption{Pattern of similar patients in the embedding space. (a) An example patient in P19 dataset. (b) The four nearest neighbors in the embedding space that are learned by TEE+DAM (b), and DAM (c). Similar patterns are highlighted in yellow.}
  \label{fig:similar_patterns}

\end{figure}

Fig. \ref{fig:similar_patterns}-a illustrates an example patient in P19 dataset. We have also visualized the four nearest neighbors in the embedding space learned by TEE+DAM and DAM models in \ref{fig:similar_patterns}-b and \ref{fig:similar_patterns}-c respectively. As can be seen, the four nearest neighbors in TEE+DAM have a much more similar pattern with respect to the example patient (some of the similar patterns are highlighted in yellow).

% This means that the learned representations are better suited to keeping patients with similar patterns closer in the embedding space. More powerful representations of electronic health records (EHRs) could have various applications in healthcare, such as synthetic data generation, and patient phenotyping.

\begin{table}[htbp]
  \centering
  \caption{10-nearest Neighbor pattern similarity (10nn-ps) for different variations of our model. mean(std) across splits are reported. The best and second-best results are highlighted in bold and underlined respectively.}

  % Table generated by Excel2LaTeX from sheet 'T4'
\begin{tabular}{lcc}
\toprule
\toprule
\multicolumn{1}{c}{Model} & P12   & P19 \\
\midrule
\textbf{DAM} & 81.2(1.4) & 67.9(2.1) \\
\textbf{TEE+DAM} & 86.5(1.0) & 73.5(1.0) \\
\textbf{TEE+DAM w AE} & 92.5(0.2) & \textbf{83.6(1.2)} \\
\textbf{TEE+DAM w PP(ML)} & 91.0(0.6) & 78.8(2.4) \\
\textbf{TEE+DAM w PP(Single)} & \textbf{92.9(0.5)} & 82.3(2.1) \\
\textbf{[TEE w AE]+DAM} & 91.7(0.9) & 82.1(3.2) \\
\textbf{[TEE w PP(ML)]+DAM} & 91.4(0.7) & 78.7(2.5) \\
\textbf{[TEE w PP(Single)]+DAM} & 91.5(1.1) & 83.3(1.6) \\
\bottomrule
\bottomrule
\end{tabular}%

  \label{tab:knn}%
\end{table}%

% TEE+DAM ($0.903 \pm 0.058 $) is significantly higher than (DAM) ($0.832 \pm 0.095$)($pvalue \leq 0.001$). The sampling pattern of an example patient and its 4 nearest neighbors (in the embedding space) is visualized in Fig. \ref{fig:knn}-c and d for TEE+DAM and TEE, respectively. As can be seen, in TEE+DAM, the 4 neighbors have a more similar pattern compared to the example patient.

% Table generated by Excel2LaTeX from sheet 'T3_new'
% \begin{table}[htbp]
%     \centering
%     \caption{10-nearest neighbour pattern similarity (10nn-ps)}

%     \import{./tables}{T4_knn_similarity.tex}

%     \label{tab:knn}%
% \end{table}%

% \begin{figure}[!h]

%     \centerline{\includegraphics{images/knn_patterns.png}}

%     \caption{Magnetization as a function of applied field.
%         It is good practice to explain the significance of the figure in the caption.}
%     \label{fig:knn}
% \end{figure}

\section{Conclusions}
\label{sec:Conclusion}

In this study, we introduce TEE4EHR, a new transformer event encoder for handling the pattern of irregularly sampled time series in EHRs based on point process theory. TEE achieves state-of-the-art performance in common event sequence datasets in neural point process literature. Furthermore, by combining TEE with an existing deep attention module, we could improve the performance metrics for outcome prediction in two real-world healthcare datasets. TEE can also achieve better patient embeddings by leveraging pattern laboratory tests. This means that the learned representations are better suited to keeping patients with similar patterns closer in the embedding space. More powerful representations of electronic health records (EHRs) could have various applications in healthcare, such as synthetic data generation, and patient phenotyping.

We also highlighted various drawbacks of current neural point process architectures and provided some guidelines for future research in this area. In the future, we plan to validate our TEE on other event sequence datasets. Another avenue of investigation involves substituting DAM with alternative architectures to explore potential performance enhancements when combined with TEE.

\bibliographystyle{unsrt}  

\bibliography{template}

\newpage

\section{Appendix A: Training Details}
We used a consistent data split for all datasets. The splits for SO and RT-MC were taken from \cite{meiNeuralHawkesProcess2017}, while the splits for RT-ML and SYN were taken from \cite{enguehardNeuralTemporalPoint2020}. The splits for P12 and P19 were taken from \cite{zhangGraphGuidedNetworkIrregularly2022}.

In P12, we selected the pattern following 24 laboratory tests to be modeled by TEE: Blood Urea Nitrogen (BUN), Creatinine, Glucose, Bicarbonate (HCO3), Sodium (Na), Potassium (K), Magnesium (Mg), Hematocrit (HCT), Platelets, White Blood Cell Count (WBC), Fraction of Inspired Oxygen (FiO2), Partial Pressure of Carbon Dioxide (PaCO2), Partial Pressure of Oxygen (PaO2), pH, Arterial Oxygen Saturation (SaO2), Alkaline Phosphatase (ALP), Alanine Aminotransferase (ALT), Aspartate Aminotransferase (AST), Albumin, Bilirubin, Lactate, Cholesterol, Troponin I, Troponin T.

For P19, we selected the following 25 laboratory tests: Blood Urea Nitrogen (BUN), Creatinine, Glucose, Bicarbonate (HCO3), Potassium, Magnesium, Hematocrit (Hct), Platelets, White Blood Cell Count (WBC), Fraction of Inspired Oxygen (FiO2), Partial Pressure of Carbon Dioxide (PaCO2), pH, Arterial Oxygen Saturation (SaO2), Alkaline Phosphatase (Alkalinephos), AST, Total Bilirubin (Bilirubin\_total), Direct Bilirubin (Bilirubin\_direct), Lactate, Troponin I, Hemoglobin (Hgb), Chloride, Phosphate, Calcium, Partial Thromboplastin Time (PTT), Fibrinogen.

We used the Adam optimizer \cite{kingmaAdamMethodStochastic2017} to optimize the model during training. Besides, we used a cosine annealing learning rate scheduler \cite{loshchilovSGDRStochasticGradient2017} which cyclically reduces the learning rate following a cosine function, allowing the model to converge faster and potentially escape local minima.

Hyperparameter tuning is performed on the validation set using Optuna library in Python. The initial value of the learning rate was 1e3, 3e-3, 3e-3, 2e-2 and 1e-2 for SO, RT, SYN, P12 and P19 respectively. Batch size is 4, 64, 64, 128, 128 for SO, RT, SYN, P12 and P19 respectively. Other model parameters can be found in the GitHub repository.

\end{document}